\documentclass[conference]{IEEEtran}
\IEEEoverridecommandlockouts

\usepackage{cite}
\usepackage{amsmath,amssymb,amsfonts}
\usepackage{algorithmic}
\usepackage{graphicx}
\usepackage{textcomp}
\usepackage{colortbl}
\usepackage{subfig} 
\usepackage{float} 
\usepackage[table]{xcolor}
\usepackage{xcolor}
\begin{document}

\def\x{{\mathbf x}}
\def\L{{\cal L}}
\newcommand{\red}{\color{red}}

\makeatother

\title{Explainable AI in Orthopedics: Challenges, Opportunities, and Prospects}

\makeatletter
\newcommand{\linebreakand}{%
  \end{@IEEEauthorhalign}
  \hfill\mbox{}\par
  \mbox{}\hfill\begin{@IEEEauthorhalign}
}


%
\makeatother
\author{
\IEEEauthorblockN{Soheyla Amirian, PhD} 
\IEEEauthorblockA{\text{\textit{University of Georgia}}\\\textit{amirian@uga.edu}}
\and \IEEEauthorblockN{Luke A. Carlson, MS} 
\IEEEauthorblockA{
\text{\textit{University of Pittsburgh}}\\\textit{lac249@pitt.edu}}
\and\IEEEauthorblockN{Matthew F. Gong, MD} 
\IEEEauthorblockA{\text{\textit{University of Pittsburgh}}\\\textit{gongm2@upmc.edu}}

\linebreakand

\and \IEEEauthorblockN{Ines Lohse, PhD} 
\IEEEauthorblockA{
\text{\textit{\textit{University of Pittsburgh}}}\\\textit{inl22@pitt.edu}}

\and\IEEEauthorblockN{Kurt R. Weiss, MD} 
\IEEEauthorblockA{\text{\textit{University of Pittsburgh}}\\\textit{krw13@pitt.edu}}
\and\IEEEauthorblockN{Johannes F. Plate, MD, PhD} 
\IEEEauthorblockA{
\text{\textit{University of Pittsburgh}}\\\textit{johannes.plate@pitt.edu}}
\and\IEEEauthorblockN{Ahmad P. Tafti, PhD} 
\IEEEauthorblockA{\text{\textit{University of Pittsburgh}}\\\textit{tafti.ahmad@pitt.edu}}
}
\maketitle

\begin{abstract}
While artificial intelligence (AI) has made many successful applications in various domains, its adoption in healthcare lags a little bit behind other high-stakes settings. Several factors contribute to this slower uptake, including regulatory frameworks, patient privacy concerns, and data heterogeneity. However, one significant challenge that impedes the implementation of AI in healthcare, particularly in orthopedics, is the lack of explainability and interpretability around AI models. Addressing the challenge of explainable AI (XAI) in orthopedics requires developing AI models and algorithms that prioritize transparency and interpretability, allowing clinicians, surgeons, and patients to understand the contributing factors behind any AI-powered predictive or descriptive models. The current contribution outlines several key challenges and opportunities that manifest in XAI in orthopedic practice. This work emphasizes the need for interdisciplinary collaborations between AI practitioners, orthopedic specialists, and regulatory entities to establish standards and guidelines for the adoption of XAI in orthopedics. 
\end{abstract}


\section{Introduction}
\label{sec:intro}
While AI has shown promise in various real-world scenarios, its adoption in healthcare, particularly in orthopedics, is hindered by its lack of explainability and interpretability. Explainability mainly refers to the ability to uncover the AI black box in a way that allows end users, including clinicians, physicians, healthcare practitioners, and patients to clearly understand how AI algorithms are making decisions. This helps the end users to trust and comprehend the reasoning behind AI systems, unlocking the full potential of AI in orthopedic care and patient and clinical outcomes \cite{mehta2022explainable}. Interpretability, similarly, refers to the ability to interpret the underlying mechanisms and features that contribute to AI-generated results \cite{mehta2022explainable}.

From a technical perspective, the taxonomy of XAI methods can be categorized into three mechanisms, including (1) local explanation, (2) global explanation, and (3) counterfactual explanation \cite{lundberg2020local, pedreschi2019meaningful,shaban2021guest, xu2019explainable}. Local explanation methods explain the decisions of AI models by focusing on individual data points. This can be helpful for understanding why a particular patient is classified in a certain way. For example, a local explanation method could show which features of a patient's EHRs were most important in leading the AI model to make a particular diagnosis (e.g., infection) \cite{lundberg2020local,pedreschi2019meaningful,amann2020explainability}. Global explanation methods explain the decisions of an AI model by looking at the model as a whole. This can be helpful for understanding how the model is making decisions and identifying potential biases in the model and data. For example, a global explanation method could illustrate which features are most important in the model's decision-making process, and whether these features are distributed evenly across different groups of patients \cite{amann2020explainability,saraswat2022explainable}. Counterfactual explanation methods generate explanations by providing alternative scenarios that would lead to different outcomes. This can be helpful for understanding how the model is sensitive and robust to different feature sets. For example, a counterfactual explanation for a patient's diagnosis could show how the diagnosis would have changed if the patient's gender, race, ethnicity, BMI, and/or blood type had been different.

There are a number of compelling reasons to demonstrate why XAI is important in orthopedics. First, XAI plays a pivotal role in establishing trust between patients and healthcare providers. When patients can understand the reasoning behind an AI model's decision regarding their care, it enhances transparency and fosters trust in the healthcare process. By engaging patients in the decision-making process, XAI empowers them to actively participate in their own healthcare management and adhere to the recommended therapy plans. Second, XAI aligns with regulatory requirements and ethical considerations in healthcare. Many regulatory frameworks, such as the General Data Protection Regulation (GDPR) in Europe \cite{regulation2018general}, emphasize the need for transparent AI and XAI systems. By incorporating XAI, healthcare organizations ensure compliance with these regulations and adhere to ethical principles, such as providing justifications for the decisions made by AI models. Third, XAI facilitates the validation and adoption of AI models in clinical practices. By providing interpretable explanations, it allows healthcare professionals to validate the model's recommendations and verify that they are consistent with their own clinical knowledge and experience. This validation process helps build confidence in the AI model's capabilities and supports its integration and uptake into existing clinical workflows. Fourth, XAI helps to identify biases by providing explanations for the model's decisions. This can make it easier to trace the factors contributing to biased outcomes, such as the data point used to train and validate the model or the way the model is built. With this knowledge, healthcare practitioners will take corrective measures to mitigate bias, ensuring fair and equitable healthcare for all patients across different gender and racial groups for example. Finally, XAI enables healthcare providers and AI developers to gain a deeper understanding of how AI models are functioning and identify areas for improvement. By unraveling the inner workings of the model, researchers can identify strengths, weaknesses, and limitations, allowing them to refine the model's performance, availability, accuracy, and reliability. 

The integration of XAI in healthcare holds significant potential for enhancing trust, mitigating bias, optimizing model performance, ensuring regulatory compliance, improving patient and clinical outcomes, facilitating clinical validation, and disseminating research findings. By implementing XAI, orthopedic centers can foster transparency, fairness, and effectiveness in the utilization of AI systems, thereby driving improved patient outcomes and advancing care. 

In this contribution, an attempt was made to outline several key challenges and opportunities of XAI in orthopedics (see figure \ref{fig:Outlines}). The organization of this work is as follows. Section II reviews the literature and highlights the recent advances in AI explainability and its applications in healthcare and orthopedics. Sections III and IV list potential opportunities and challenges in this domain, respectively. Further discussions are presented in Section V.

\vspace{.1cm}
\begin{figure}
\centering
    \includegraphics[width=\linewidth]
    {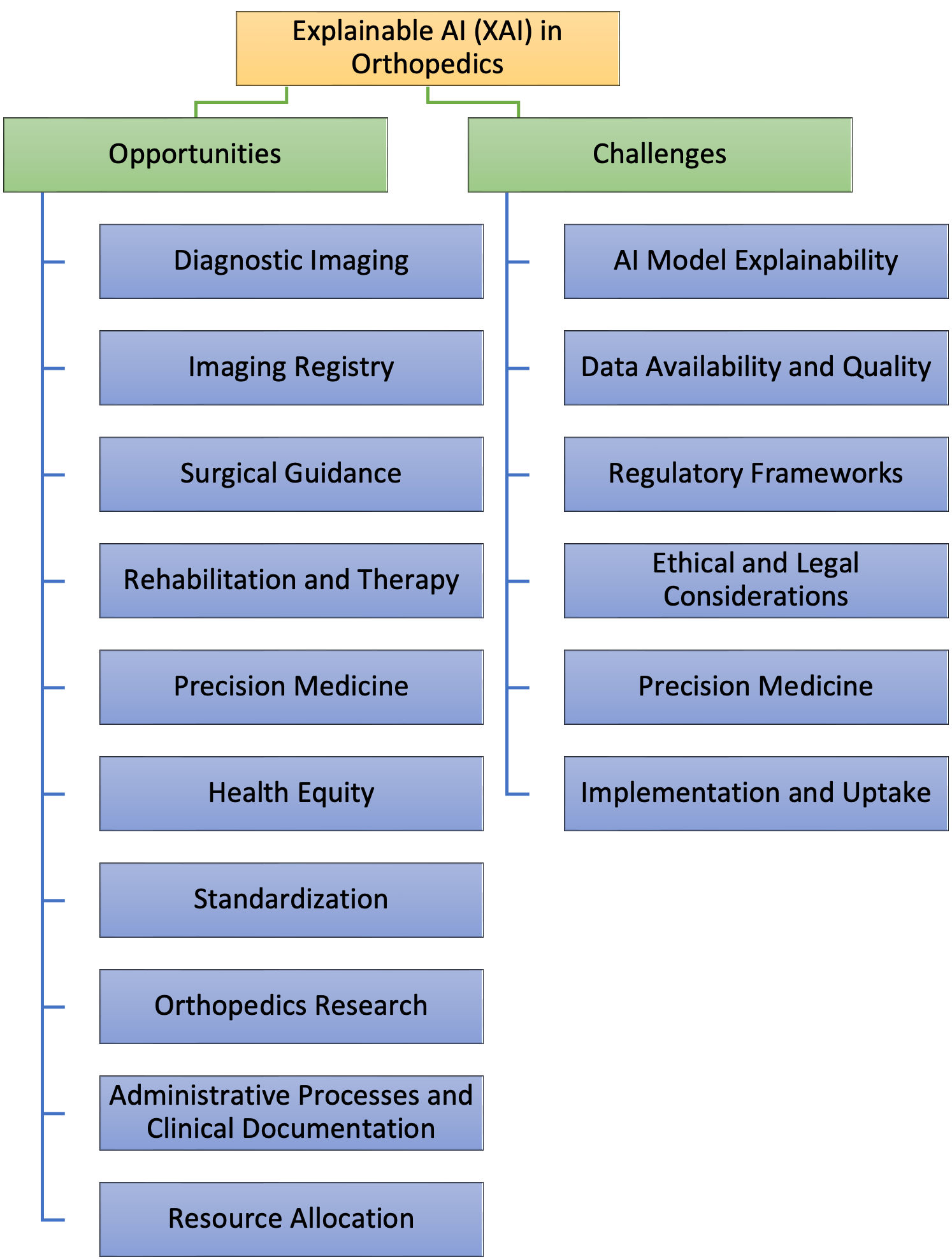}
    \caption{Taxonomy of opportunities and challenges of explainable AI (XAI) in orthopedics.} 
    \label{fig:Outlines}
\end{figure}
\vspace{.1cm}
\section{Background}
\label{sec:bkgd}

In this section, we explore the burgeoning field of XAI. With the rapid progress and growing interest in XAI, it has become imperative to provide thorough overviews of the latest advancements in state-of-the-art XAI. To address this need, we meticulously examined a selection of research papers that delve into the intricacies of XAI, shedding light on the methodologies, techniques, and applications in various domains.

Karim et al. \cite{karim2021deepkneeexplainer} presented DeepKneeExplainer, a novel method for explainable knee osteoarthritis (OA) diagnosis using radiographs and MRIs. Through experiments on multicenter osteoarthritis study (MOST) cohorts, their approach demonstrates remarkable classification accuracy, outperforming comparable state-of-the-art techniques. The use of deep-stacked transformation chains ensures robustness against potential noise and artifacts in test cohorts. Additionally, WeightWatcher is applied to address model selection bias. The authors proposed integrating this approach into clinical settings to enhance domain generalization for biomedical images. Although promising, further clinical experiments and improvements are necessary. The paper encourages the adoption of explainable methods and DNN-based analytic pipelines in clinical practice to promote AI-assisted applications. 
Subsequently, Kokkotis et al.
\cite{kokkotis2022explainable} focused on Knee Osteoarthritis (KOA), aiming to reduce diagnostic errors by providing reliable tools for diagnosis. Using multidimensional data from the Osteoarthritis Initiative database, the researchers proposed a robust Feature Selection (FS) methodology based on fuzzy logic.
They emphasized the need for explainability analysis using SHAP to understand the model's decision-making process and the impact of selected features. Overall, the proposed methodology offered an approach for identifying informative risk factors in KOA diagnosis, with potential applications in other medical domains.

Mohseni et al. \cite{mohseni2021multidisciplinary} presented a comprehensive survey and framework aimed at facilitating the sharing of knowledge and experiences in the design and evaluation of XAI systems across various disciplines. The framework provides a categorization of XAI design goals and evaluation methods, mapping them to different user groups of XAI systems. They proposed a step-by-step design process and evaluation guidelines to assist multidisciplinary XAI teams in their iterative design and evaluation cycles. They emphasize the importance of using appropriate evaluation measures for different design goals and advocate for a balance between qualitative and quantitative methods during the design process. The framework addresses the overlap of XAI goals among different research disciplines and highlights the significance of considering user interactions and long-term evaluation in XAI system design.
Wen Loh et al.
\cite{loh2022application} discussed the importance of XAI in healthcare to build trust in AI models and encourage their practical use. They focused on various XAI techniques used in healthcare applications and surveyed 99 articles from highly credible journals (Q1) covering techniques such as SHAP, LIME, GradCAM, LRP, and others. They identified areas in healthcare that require more attention from the XAI research community, specifically detecting abnormalities in 1D biosignals and identifying key text in clinical notes. The most widely used XAI techniques are SHAP for explaining clinical features and GradCAM for providing visual explanations of medical images. The work concluded that a holistic cloud system for smart cities can significantly advance healthcare by promoting the use and improvement of XAI in the industry.

As the application of AI continues to expand, particularly in critical domains like healthcare and medical imaging diagnosis, the demand for interpretable and transparent AI models becomes paramount.
Patrício et al.
\cite{patricio2022explainable} provided a comprehensive survey of XAI applied to medical imaging diagnosis. They addressed the lack of interpretability in deep learning models and the need for XAI to explain the decision-making process. Their study covered various XAI techniques, including visual, textual, example-based, and concept-based methods. They also discussed existing medical imaging datasets and evaluation metrics for explanations. They emphasized the importance of inherently interpretable models and textual explanations. Challenges in medical image interpretability were identified, including the need for larger datasets and the use of self-supervised learning. They highlighted the importance of objective metrics for evaluating explanations and the potential of using Transformers for report generation in medical imaging.

Agarwal et al.
\cite{agarwal2022openxai} introduced OpenXAI.
OpenXAI is an open-source framework designed to evaluate and benchmark post hoc explanation methods systematically. It provides a collection of real-world datasets, pre-trained models, and state-of-the-art feature attribution methods, along with twenty two quantitative metrics to assess the faithfulness, stability, and fairness of these methods. The framework includes XAI leaderboards, enabling easy comparison of explanation methods across various datasets and models. OpenXAI promotes transparency, reproducibility, and standardization in benchmarking and simplifying the evaluation process for researchers and practitioners. It supports extensibility, allowing users to incorporate custom methods and datasets. The framework aims to ensure reliable post hoc explanations for decision-makers in critical applications.

As a case study, we reference the work of Littlefield et al. \cite{nick2023}, which introduced an explainable deep few-shot learning model. This model successfully identified and delineated the knee joint area in plain knee radiographs by utilizing a limited number of manually annotated radiographs.

In this section, we provided a comprehensive exploration of XAI and its applications in various domains, particularly in medical imaging diagnosis and healthcare. The reviewed papers highlighted the importance of XAI in enhancing model interpretability, promoting trust, and encouraging the practical use of AI systems. 
The advancements in XAI showcased in the reviewed papers demonstrate the growing interest and potential of this field in improving decision-making processes and facilitating AI adoption in critical applications.
\section{Opportunities}
\label{sec:Opp}
The incorporation of XAI in orthopedics has several benefits. Here, we discuss potential opportunities for using XAI in orthopedics.

\subsection{Diagnostic Imaging} 
AI models have provided promising results in atrribution-based diagnostics of orthopedic conditions through the analysis of medical images, including X-rays, CT scans, and MRIs. These models are trained and validated on extensive imaging datasets and patient data, enabling them to learn and recognize patterns associated with various complications, such as bone lucency, bone loosening, and bone deformity. By offering interpretable justifications for AI decisions, XAI can assist in making these models more transparent and intelligible. These methods aid in bridging the comprehension gap between model output and end users by offering explanations, such as identifying the precise region of interest (ROI) or elements of the image and/or imaging biomarkers that contributed to the diagnosis. This can facilitate more informed and shared decision-making about patient care by assisting healthcare providers, patients, and physicians in better understanding of the rationale behind AI models \cite{miller2022explainable, gulum2021review, singh2020explainable}.

\subsection{Imaging Registry}
XAI is of paramount importance in establishing an imaging registry in orthopedics by providing explainable analysis of medical imaging, standardized data analysis, and making the data more accessible and usable. XIA can help with automatically and objectively extracting relevant information from imaging data, identifying patterns, and categorizing various orthopedic complications in a consistent manner, which will ensure uniformity and reliability in the analysis of imaging data within the imaging registry \cite{nazir2023survey, yang2022unbox, bourdon2021explainable}. Moreover, XAI algorithms can also be used to detect and correct errors in imaging data, ensuring that the data is accurate and consistent.

\subsection{Surgical Guidance} 
XAI can assist surgical guidance in orthopedics by providing surgeons with interpretable explanations for the decisions made by AI-powered surgical guidance systems. This capability enables surgeons to comprehend the underlying rationale behind the system's recommendations, fostering trust and ultimately leading to improved surgical outcomes. For instance, XAI may be used to explain why a surgical strategy is recommended by an AI-powered model. A summary of the considerations that went into a surgeon's choice, such as the patient's medical history and imaging biomarkers, might be given via the system. This would allow surgeons to understand the system's reasoning, making informed decisions about whether to follow AI recommendations  \cite{zhang2022applications, o2022explainable}. Furthermore, XAI could be used to identify potential risks associated with a surgical plan. The system could highlight areas of the patient's anatomy that are at risk of injury or it could identify potential complications (e.g., fracture) that could arise from the surgery. Presumably, this would allow surgeons to take steps to better reduce these risks and to improve the safety of the procedure.

\subsection{Rehabilitation and Therapy} 
For orthopedic patients, XAI can be used to create individualized rehabilitation and therapy plans. XAI can help therapists customize treatment plans and track the success of interventions by offering interpretable insights into the patient's progress, areas that need improvement, and suggested exercises. This enables personalized treatment plans, open progress monitoring, increased patient engagement, and decision support for therapists. Therapists can maximize treatment results and raise the standard of care given to those patients by implementing XAI into orthopedic rehabilitation techniques \cite{de2023implementation, gandolfi2022explainable, misic2022overview}. 

\subsection{Precision Medicine} 
By providing explanations regarding the contributions of discrete variables, including patient demographics, patient medical history, EHRs, and imaging data, XAI empowers orthopedic specialists in understanding the reasons behind treatment recommendations, considering various patient-specific factors, thus it assists orthopedic surgeons in creating personalized treatment plans tailored to individual patients, optimizing the chances of successful patient outcomes \cite{kamel2021digital, gimeno2022explainable}. By augmenting the physician's expertise with transparent explanations, XAI helps toward more informed and confident diagnosis and treatment decisions, ultimately leading to improved patient and clinical outcomes.

\subsection{Health Equity} 
In orthopedics, XAI can help discover and treat health inequities. XAI can assist in identifying systemic problems, socioeconomic determinants of health, and obstacles to fair care by utilizing large-scale datasets and generating explanations for differences in treatment outcomes across various populations \cite{berdahl2023strategies}. This knowledge can inform interventions and policies aimed at reducing disparities and improving health equity, ensuring that decisions are made with fairness, equity, and optimal patient well-being \cite{arrieta2020explainable, alikhademi2021can, rai2020explainable}.

\subsection{Standardization} 
XAI can improve the standardization and consistency of orthopedic care. By clarifying pre- and post-operative recommendations, XAI helps ensure that consistent decisions are made for similar cases  regardless of  healthcare provider, computer system, or facility. By explaining the rationale behind treatment recommendations,  XAI models  share information and decision-making processes gathered by leading experts. This helps disseminate best practices and standardize care across healthcare settings, minimizing unnecessary variation in care and ensuring consistent access to care regardless of provider expertise \cite{albahri2023systematic, cina2022we, bharati2023review}.

\subsection{Orthopedics Research}
XAI contributes to orthopedic research by analyzing large-scale datasets and scientific literature, identifying patterns, and generating explanations for observed correlations. This will help researchers in uncovering new insights, understanding disease mechanisms, treatment strategies, and discovering potential risk factors or treatment approaches.

\subsection{Administrative Processes and Clinical Documentation} 
XAI can be used to automate administrative tasks, improve the accuracy and quality of clinical documentation, and optimize administrative workflows. Using AI with explainable capabilities, XAI systems can handle routine administrative processes, freeing up valuable time for healthcare professionals to focus on more critical patient care tasks. They can also provide  real-time recommendations and clarifications to clinicians, promoting completeness, accuracy, and adherence to coding and documentation guidelines. In addition, XAI models generate explanations for the underlying factors that affect workflow efficiency, enabling healthcare managers to identify areas for improvement, streamline processes and allocate resources efficiently \cite{tjoa2020survey, panigutti2023co, moradi2022deep, ho2022explainability}.

\subsection{Resource Allocation} 
XAI assists in resource allocation and capacity planning by analyzing large-scale historical data and generating explanations for resource utilization patterns. This helps healthcare administrators make precise decisions about staffing levels, beds, equipment allocation, and facility utilization, ensuring efficient use of resources and maintaining optimal patient care standards \cite{chen2020using}. XAI can also be utilized to analyze data on patient wait times and to identify areas where wait times can be reduced. This information can then be used to make changes to the clinical workflow or to reallocate resources.

\section{Challenges}
\label{sec:Cha}
This section discusses the most notable challenges that can be a barrier to the adoption of XAI in orthopedics.

\subsection{AI Model Explainability}
AI explainability may come as a challenge by itself. For example, advanced deep neural network models can be difficult to explain. Extracting explanations and interpretation from those advanced AI models poses a challenge, as their internal workings may involve numerous hyper-parameters, several layers, and intricate interactions, which can make it harder for orthopedic specialists to understand how the models make decisions \cite{saraswat2022explainable, nazar2021systematic, amann2020explainability}. On the other hand, there is often a trade-off between the interpretability, transparency, and performance of AI models. More intuitive models, such as decision trees or rule-based systems, may sacrifice some predictive accuracy and precision compared to complex models. In healthcare, where accurate predictions are important, but understanding the rationale behind  predictions is equally important, striking the right balance between explainability and model performance is critical. It should be emphasized here that the development of methods for accurate interpretation of complex AI models in orthopedics is still an active  area of research.

\subsection{Data Availability and Quality}
The quality, quantity, and availability of data is a major challenge for XAI in orthopedics. Orthopedic data is often complex, very large-scale, and heterogeneous, and it can be difficult to obtain accurate and complete data. Ensuring diverse, unbiased, and high-quality data is a critical challenge to train, validate, and test XAI systems for orthopedic healthcare. 

\subsection{Regulatory Frameworks}
Regulatory frameworks and policy-makers may not yet have established clear guidelines or standards specifically addressing XAI in healthcare \cite{ghassemi2021false}, including orthopedics. Since the use of XAI in healthcare is poorly regulated by now, it will be difficult to develop and deploy XAI models in orthopedics. Even existing regulations, such as data protection and privacy, may need to be applied to the unique context of XAI in orthopedics. 
Furthermore, different countries may have disparate regulatory frameworks governing AI in healthcare. This can make it difficult for organizations operating across borders, as they have to comply with multiple regulations, which can lead to uneven enforcement. This will certainly create uncertainty for healthcare organizations and developers about compliance requirements and best practices.

\subsection{Ethical and Legal Considerations}
Implementing XAI in orthopedics requires addressing a number of ethical and legal considerations, such as patient privacy, informed consent, and the potential for unintended consequences or misuse of AI technology \cite{albahri2023systematic, ploug2020four, adadi2020explainable}. For example, XAI models often utilize sensitive patient data, such as medical images, EHRs, and clinical notes. It is important to ensure that this data is protected and that patients' privacy is fully respected. Additionally, XAI models can be complex and it is important to be aware of the potential for unintended consequences or misuse. For example, these AI models could be used to discriminate against certain patient populations or to make biased decisions.

\subsection{Precision Medicine}
While XAI can advance personalized medicine and care, AI models trained on very large-scale datasets can struggle to provide personalized explanations for individual patients. Each patient's unique characteristics, demographics, medical history, and comorbidities may not fully overlap with the training data, making it difficult to develop meaningful explanations at the instance level. This is because AI models are trained on large datasets of patient data, but this data may not be representative of all patients. For example, the training data may not include patients with rare diseases (e.g., Osteosarcoma), or patients from different cultures or marginal groups. As a result, the AI model may not be able to make accurate predictions or provide relevant explanations for some patients.

\subsection{Implementation and Uptake}
Integrating XAI into the current clinical workflow presents a significant challenge. Healthcare providers ask for user-friendly interfaces and toolsets that deliver explanations in a clear and comprehensible fashion, without imposing additional complexity on their already demanding daily routines. The integration process entails addressing multiple challenges, including developing intuitive interfaces, providing appropriate training to healthcare professionals, and ensuring the availability of necessary resources for successful implementation and long-term maintenance of the technology.

\section{Discussion, Conclusion, and Outlook}
\label{sec:Conclusion}
In orthopedics, artificial intelligence (AI) is expanding quickly and has a lot of potential to help with diagnosis, prognosis, therapy, and rehabilitation. Explainable AI (XAI), which offers justifications and interpretations for AI models, is now one of the most promising areas of AI research in orthopedics. This is crucial for orthopedics because it enables surgeons, medical professionals, stakeholders, and -most importantly- patients to comprehend how AI-powered mechanisms make decisions, and to trust the AI's predictive and descriptive results.

In this paper, we have identified the most notable opportunities and challenges associated with implementing XAI in orthopedics. First, we believe multidisciplinary collaboration is essential for the successful uptake of XAI in orthopedics. Healthcare professionals, researchers, AI scientists, industry stakeholders, policy-makers, and regulatory entities should work collaboratively to address challenges, share expertise, and implement best practices, to mainly develop clear and adaptable frameworks that address the unique aspects of XAI in orthopedics. In this way, healthcare professionals and industry stakeholders will provide insights into the clinical needs of patients and the challenges of using XAI in clinical settings, while policy-makers will establish policies that support the utilization and development of XAI in healthcare, and in particular in orthopedics. On the other side, AI scientists will build XAI-powered systems that are trustworthy, reliable, and explainable, and regulatory entities will be developing guidelines and instructions for the use of XAI in orthopedics. To promote the adoption of XAI in orthopedics, it is essential to illustrate its value to stakeholders. This can be done by showcasing successful use cases and providing evidence of improved diagnostic and prognostic accuracy, treatment outcomes, and patient satisfaction resulting from the adoption of XAI.

User-centric design methodologies will be essential to ensure seamless integration of XAI systems into the workflow of healthcare providers. By involving healthcare providers and end-users, such as physicians and patients in the design process, it will be possible to establish computational frameworks that meet their needs, provide clear and actionable explanations, and enhance shared decision-making. To achieve this, we recommend an incremental approach, starting with small-scale pilot projects, to first assess the feasibility, effectiveness, and user acceptance of XAI in specific orthopedic settings (e.g., pain progression analysis using multi-modal data). Then we recommend regularly collecting feedback from end-users and addressing identified issues or concerns to make continuous improvements. Gradually, by expanding XAI implementation based on lessons learned, we will ensure a well-organized transition and a higher success rate. We also believe in comprehensive and continuous training for healthcare professionals adopting XAI. Equipping them with a clear understanding of the XAI concepts, benefits, and limitations will help them to effectively interpret and utilize the explanations generated automatically and objectively by AI models. This training builds confidence, promotes acceptance, and facilitates the successful integration of XAI into clinical practice.

Furthermore, the safe and ethical deployment of XAI in orthopedic environments requires careful consideration. This entails adhering to patient privacy and data protection laws, continuous supervision, ongoing monitoring, ensuring equality and fairness in AI models, valuing patient autonomy and informed consent, giving patients the confidence to make informed health choices, and setting up principles and procedures for ethical XAI usage. 

In conclusion, XAI is rapidly transforming the healthcare industry, and orthopedics is no exception. XAI-powered tool sets have the potential to improve patient and clinical outcomes in a number of ways. While implementing XAI in orthopedics presents various opportunities and challenges, adopting a collaborative approach, emphasizing user-centric methodology, offering comprehensive training, addressing ethical considerations, highlighting value, conducting small-scale pilot projects, and enabling continuous evaluation and monitoring are key components for success. Our future work focuses on developing advanced techniques for AI explainability in orthopedics, including exploring novel AI explainability methods for interpreting complex AI models, enhancing the transparency of deep learning-powered algorithms, and designing visualization strategies to provide more actionable explanations in different orthopedic settings, benchmarking AI explanation methods using evaluation metrics.

\section*{Acknowledgment}

The authors declare that they have no competing interests. 

\bibliographystyle{IEEEbib}
\bibliography{refs}

\end{document}